\documentclass[a4paper,twoside]{article}

\usepackage{epsfig}
\usepackage{subcaption}
\usepackage{calc}
\usepackage{amssymb}
\usepackage{amstext}
\usepackage{amsmath}
\usepackage{amsthm}
\usepackage{multicol}
\usepackage{pslatex}
\usepackage{apalike}
\usepackage{hyperref}

\usepackage{graphicx}
\usepackage{caption}
\usepackage{xspace}
\usepackage{enumitem}

\let\bibhang\relax
\usepackage{natbib}

\newcommand{\figref}[1]{Figure~\ref{#1}} %

\newcommand{\tabref}[1]{Table~\ref{#1}} %

\newcommand{\secref}[1]{Section~\ref{#1}}

\makeatletter

\DeclareRobustCommand\onedot{\futurelet\@let@token\@onedot}

\newcommand\@onedot{\ifx\@let@token.\else.\null\fi\xspace} %

\newcommand\eg{\emph{e.g}\onedot} %

\newcommand\ie{\emph{i.e}\onedot} %

\newcommand\cf{\emph{cf}\onedot} %

\newcommand\wrt{w.r.t\onedot} %

\newcommand\etal{\emph{et al}\@onedot} %

\makeatother

\DeclareMathAlphabet{\mathcal}{OMS}{cmsy}{m}{n}
\SetMathAlphabet{\mathcal}{bold}{OMS}{cmsy}{b}{n}

\usepackage{mathrsfs}
\usepackage{SCITEPRESS}     %

\begin{document}

\title{Improving Semantic Image Segmentation via Label Fusion in Semantically Textured Meshes}

\author{
	\authorname{Florian Fervers, Timo Breuer, Gregor Stachowiak, Sebastian Bullinger, Christoph Bodensteiner, Michael Arens}
	\affiliation{Fraunhofer IOSB, 76275 Ettlingen, Germany}
	\email{\{f\_author, s\_author\}@iosb.fraunhofer.de}
}


\abstract{Models for semantic segmentation require a large amount of hand-labeled training data which is costly and time-consuming to produce. For this purpose, we present a label fusion framework that is capable of improving semantic pixel labels of video sequences in an unsupervised manner. We make use of a 3D mesh representation of the environment and fuse the predictions of different frames into a consistent representation using semantic mesh textures. Rendering the semantic mesh using the original intrinsic and extrinsic camera parameters yields a set of improved semantic segmentation images. Due to our optimized CUDA implementation, we are able to exploit the entire $c$-dimensional probability distribution of annotations over $c$ classes in an uncertainty-aware manner. We evaluate our method on the Scannet dataset where we improve annotations produced by the state-of-the-art segmentation network ESANet from $52.05 \%$ to $58.25 \%$ pixel accuracy. We publish the source code of our framework online to foster future research in this area (\texttt{\url{https://github.com/fferflo/semantic-meshes}}). To the best of our knowledge, this is the first publicly available label fusion framework for semantic image segmentation based on meshes with semantic textures.}

\onecolumn \maketitle \normalsize \setcounter{footnote}{0} \vfill

\section{\uppercase{Introduction}}
\label{sec:introduction}

Semantic image segmentation plays an important role in computer vision tasks by providing a high-level understanding of observed scenes. However, good segmentation results are limited by the quality and quantity of the available training data which requires a lot of time-consuming manual annotation work. The popular Cityscapes dataset of traffic scenes for example reports upwards of 90 minutes per image for pixel-wise annotations \citep{Cordts2016Cityscapes}.

Since labeled datasets are rare and often cover only narrow use cases, we consider the unsupervised enhancement of predicted image segmentations of video sequences by defining consistency constraints that reflect temporal and spatial structure properties of the captured scenes.

\begin{figure}[t]
	\begin{subfigure}[t]{0.235\textwidth}
		\centering
		\includegraphics[height=27mm]{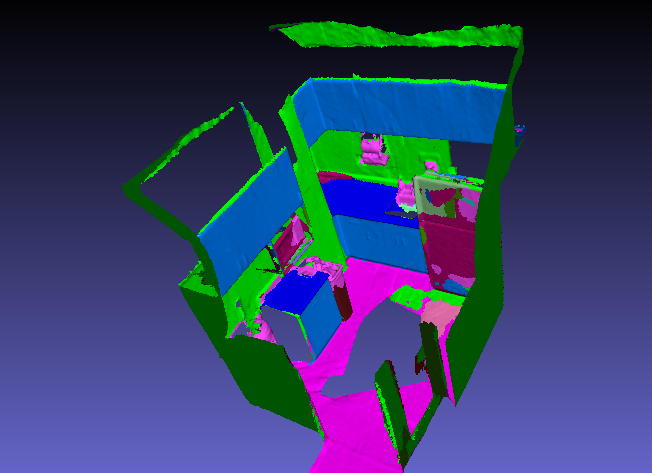}
	\end{subfigure}%
	\begin{subfigure}[t]{0.235\textwidth}
		\centering
		\includegraphics[height=27mm]{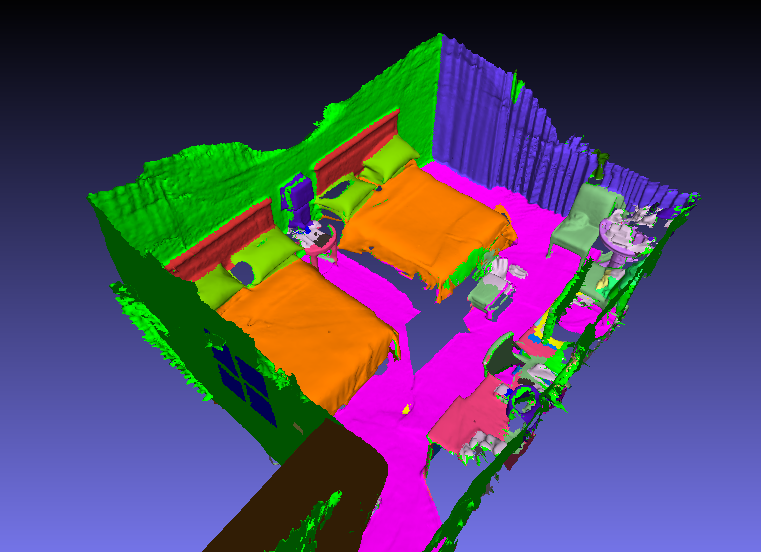}
	\end{subfigure}
	
	\vspace{-0.5mm}\begin{subfigure}[t]{0.235\textwidth}
		\centering
		\includegraphics[height=27mm]{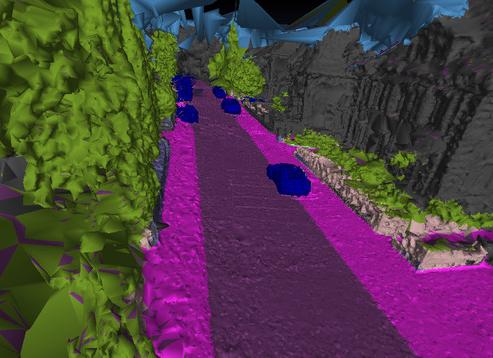}
	\end{subfigure}%
	\begin{subfigure}[t]{0.235\textwidth}
		\centering
		\includegraphics[height=27mm]{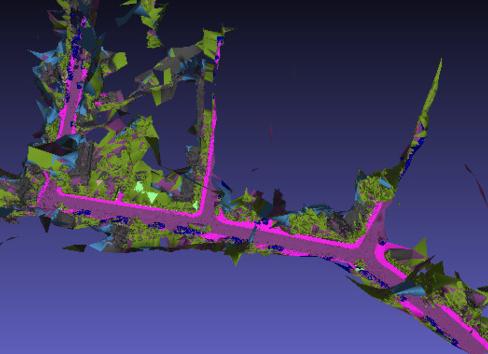}
	\end{subfigure}
	
	\caption{Semantically textured meshes of indoor and outdoor scenes produced by our label fusion framework and visualized with MeshLab \citep{meshlab}.}
	\label{fig:overview}
\end{figure}

The majority of recent works in this area has focused on methods that establish \textit{short-term} pixel correspondences -  for example via optical flow \citep{gadde2017semantic,mustikovela2016can,nilsson2018semantic}, patch match \citep{badrinarayanan2010label,budvytis2017large}, learned correspondences \citep{zhu2019improving} or depth and relative camera pose of subsequent frames \citep{ma2017multi,stekovic2020casting}. In this work, we explore a different approach by explicitly modeling the environment as a 3D semantically textured mesh which serves as a \textit{long-term} temporal and spatial consistency constraint.

\newcommand\figwidth{0.236}
\newcommand\figheight{28mm}
\begin{figure*}[t]
	\centering
	\begin{subfigure}[t]{\figwidth\textwidth}
		\includegraphics[height=\figheight]{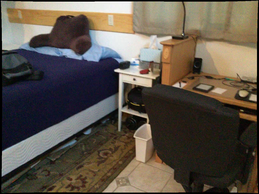}
	\end{subfigure}%
	\begin{subfigure}[t]{\figwidth\textwidth}
		\includegraphics[height=\figheight]{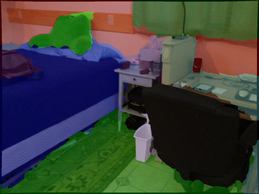}
	\end{subfigure}%
	\begin{subfigure}[t]{\figwidth\textwidth}
		\includegraphics[height=\figheight]{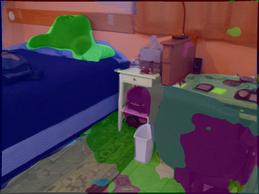}
	\end{subfigure}%
	\begin{subfigure}[t]{\figwidth\textwidth}
		\includegraphics[height=\figheight]{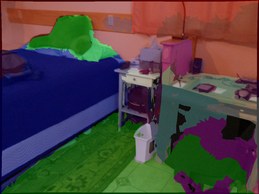}
	\end{subfigure}
	
	\vspace{-0.2mm}
	
	\begin{subfigure}[t]{\figwidth\textwidth}
		\includegraphics[height=\figheight]{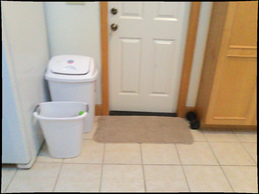}
	\end{subfigure}%
	\begin{subfigure}[t]{\figwidth\textwidth}
		\includegraphics[height=\figheight]{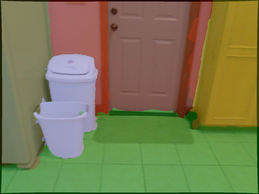}
	\end{subfigure}%
	\begin{subfigure}[t]{\figwidth\textwidth}
		\includegraphics[height=\figheight]{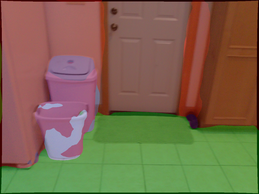}
	\end{subfigure}%
	\begin{subfigure}[t]{\figwidth\textwidth}
		\includegraphics[height=\figheight]{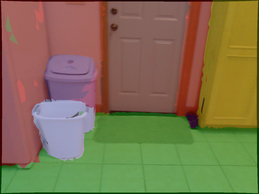}
	\end{subfigure}

	\begin{subfigure}[t]{\figwidth\textwidth}
		\includegraphics[height=\figheight]{images/new/900_color}
		\caption{Color}
	\end{subfigure}%
	\begin{subfigure}[t]{\figwidth\textwidth}
		\includegraphics[height=\figheight]{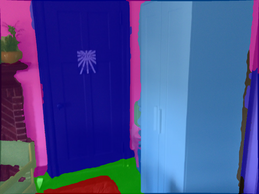}
		\caption{Ground truth}
	\end{subfigure}%
	\begin{subfigure}[t]{\figwidth\textwidth}
		\includegraphics[height=\figheight]{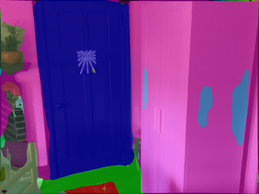}
		\caption{Network prediction}
	\end{subfigure}%
	\begin{subfigure}[t]{\figwidth\textwidth}
		\includegraphics[height=\figheight]{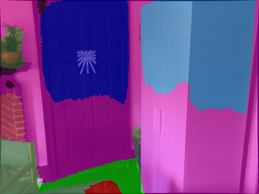}
		\caption{Fused annotation}
	\end{subfigure}
	
	\caption{Results of the label fusion in the Scannet dataset. The fusion shows most improvement when seeing an object from many different perspectives. Errors in the reconstructed mesh result in artifacts in the rendered annotations. The figure is best viewed in color. \label{fig:qualitative}}
	\vspace{-0.3cm}
\end{figure*}

The main contributions of this work are as follows:\\

\renewcommand*\labelenumi{(\theenumi)}
\begin{enumerate}[wide, labelwidth=!, labelindent=0pt]
	\item We present a label fusion framework based on environment mesh reconstructions that is capable of improving the quality of semantic pixel-level labels in an unsupervised manner. In contrast to previous works, we show that the proposed method improves segmentation results even without requiring depth sensors for the mesh reconstruction.
	\item We introduce a novel pixel-weighting scheme that dynamically adjusts the contribution of individual frames towards the final annotations. This yields a larger improvement in pixel accuracy than previous label fusion works.
	\item Our framework uses a custom renderer and texture parametrization to optimize GPU memory utilization during the label fusion process. This allows us to exploit the entire $c$-dimensional probability distribution of annotations over $c$ classes in an uncertainty-aware manner. We implement the entire framework in CUDA \citep{nickolls2008scalable}, since the texture mapping techniques of classical rendering pipelines such as OpenGL \citep{woo1999opengl} that are used in previous works are not suited for this task.
	\item We make the code to our framework publicly available, including all evaluation scripts used for this paper.
\end{enumerate}

\section{\uppercase{RELATED WORK}}

Short-term correspondences between pixels of subsequent video frames have been widely used to improve semantic segmentation \citep{gadde2017semantic,mustikovela2016can,nilsson2018semantic,badrinarayanan2010label,budvytis2017large,zhu2019improving,ma2017multi,stekovic2020casting}. For example, \citet{zhu2019improving} and \citet{mustikovela2016can} propagate labels from hand-annotated video frames to adjacent (unlabeled) frames as data augmentation for training a segmentation model. Nevertheless, these methods are limited to short-term correspondences since label accuracy decreases with each propagation step \citep{mustikovela2016can}.

To establish long-term correspondences, most works explicitly represent the environment as a three-dimensional model and find corresponding pixels via their model projections. Voxel maps \citep{kundu2014joint,stuckler2015dense,li2017fast,grinvald2019volumetric,rosinol2020kimera,jeon2018semantic,pham2019real} have been used for this purpose, but suffer from discretization and high memory requirement. Point clouds \citep{floros2012joint,hermans2014dense,li2016semi,tateno2017cnn} on the other hand accurately represent temporal but not spatial correspondences and only model the environment in a sparse or semi-dense way. \citet{mccormac2017semanticfusion} use surfels as a dense representation of the environment and model the class probability distribution of each surfel to fuse the two-dimensional network predictions from different view points.
Similar to our work, \citet{rosu2020semi} use a semantically enriched mesh representation of the environment to fuse the predictions of individual frames. In contrast to their work, we show that our framework is able to improve segmentations even with an off-the-shelf image-based reconstruction pipeline. Additionally, \citet{rosu2020semi} use OpenGL for rendering the semantic mesh and therefore have to resort to a classical texture mapping technique that maps the entire mesh onto a single texture. We implement our framework in CUDA which enables the use of a custom texture parametrization and avoids expensive paging operations when interacting with OpenGL textures. While \citet{rosu2020semi} only use the maximum of the probability distribution, we are able to exploit the entire $c$-dimensional probability distribution of pixel labels over $c$ classes.

\section{\uppercase{METHOD}}

Our method is designed to fuse temporally and spatially inconsistent pixel predictions, \eg of a segmentation network, into a consistent representation in an uncertainty-aware manner. For this purpose, we first establish correspondences between image pixels that are projections of the same three-dimensional environment primitive by using the intrinsic and extrinsic camera parameters. The predicted class probability distributions of corresponding pixels are then aggregated resulting in a single probability distribution for the primitive. Finally, the primitive's fused annotation is rendered onto all corresponding pixels to produce consistent 2D annotation images.

\subsection{Environment Mesh}
\label{sec:environment-mesh}

To determine the three-dimensional environment primitives, a mesh of the scene captured with a set of images is reconstructed using off-the-shelf reconstruction frameworks like BundleFusion \citep{dai2017bundlefusion}, if depth information is available, and Colmap \citep{schoenberger2016sfm,schoenberger2016mvs}, when only working with RGB input. This recovers static parts of the environment as well as intrinsic and extrinsic camera parameters of individual frames. Pixels are then defined as correspondences if they stem from projections of the same primitive element in the environment mesh.

Since the reconstruction step does not explicitly account for semantics, the geometric borders of mesh triangles and the semantic borders of real-world objects are not guaranteed to always coincide. Individual triangles might therefore span across multiple semantic objects and lead to incorrect pixel correspondences. This problem can be alleviated by increasing the resolution of the mesh to include sub-triangle primitive elements with smaller spatial extension. This reduces the total number of pixel correspondences during the label fusion, but at the same time decreases the proportion of incorrect correspondences. To increase the resolution while preserving the geometry of the mesh, we introduce semantic textures that further subdivide a triangle into smaller texture elements called \textit{texels} \citep{glassner1989introduction}.

\begin{figure}
	\centering
	\includegraphics[width=40mm]{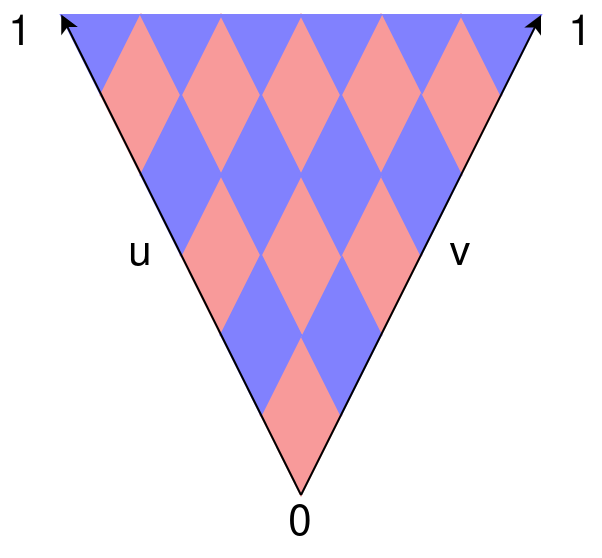}
	\caption{Example texture mapping of a single triangle. $uv$ dimensions are discretized into $s=6$ steps each yielding $\frac{s^2 + s}{2}=21$ texels.}
	\label{fig:texels}
\end{figure}

Let $t \in T$ be a triangle in mesh $T$ and $X_t$ the set of texels on the triangle. We choose a $uv$ representation for texture coordinates on the triangle \citep{heckbert1989fundamentals} with the $uv$ coordinates $(0, 0)$, $(0, 1)$ and $(1, 0)$ for the three vertices as well as $u \ge 0$, $v \ge 0$ and $u + v < 1$ for any point on the triangle. The $uv$ space is discretized into $s \in \mathbb{N}$ steps per dimension yielding a total of $|X_t|=\frac{s^2 + s}{2}$ texels (\cf \figref{fig:texels}). A texel $x \in X_t$ covers the subspace $[u_x, u_x + \frac{1}{s}) \times [v_x, v_x + \frac{1}{s})$ of the triangle's $uv$ space. To reduce the skew of texel shapes we also choose the vertex which is located on the triangle's interior angle closest to a right angle as origin of the $uv$ space.

We choose $s = \max (1, \lceil \gamma \sqrt{a_t} \rceil)$ based on the worst-case frame in which the triangle occupies the most number of pixels $a_t$. We take the square root of the area $a$ so that $|X_t| \in \mathcal{O}(a_t)$. The variable $\gamma$ serves as a tunable parameter defining the resolution of the mesh, such that for a given triangle larger $\gamma$ lead to more texels and smaller $\gamma$ to fewer texels. At $\gamma = 0$ triangles are not subdivided into texels. This formulation of texture resolution aims to be agnostic \wrt the granularity of the input mesh: For a given $\gamma > 0$ texels will roughly encompass the same number of pixels on the worst-case frame regardless of the triangle size.

Let $K$ further denote the set of all pixels in all images. For a pixel $k \in K$, let $t_k$ denote the projected triangle and $(u_k, v_k)$ the corresponding $uv$ coordinates. We define $K_x \subset K$ as the set of pixels that are projected onto a given texel $x$ according to \eqref{eq:projected-pixels}, and $x_k$ as the texel that pixel $k$ is projected onto, such that $k \in K_{x_k}$.

\begin{multline}
	\label{eq:projected-pixels}
	K_x = \{k | t_k = t_x \\
	\land (u_k, v_k) \in [u_x, u_x + \frac{1}{s}) \times [v_x, v_x + \frac{1}{s}) \}
\end{multline}

\subsection{Label Fusion}

In general, a single texel is projected onto a set of pixels in multiple images where each pixel $k$ is annotated with a probability distribution $\mathbf{p}_k \in \mathbb{R}^c$ over $c$ possible classes. Each pixel prediction represents an observation of the underlying texel class. Potentially conflicting pixel predictions are fused in an uncertainty-aware manner into a single probability distribution $\mathbf{p}_x\in \mathbb{R}^c$ that represents the label of the texel $x$.

\subsubsection{Aggregation}
\label{sec:theory-aggregation}

Let $f: \mathcal{P}(K) \mapsto \mathbb{R}^c$ represent an aggregation function that maps a set of pixels onto a fused probability distribution. The annotation $\mathbf{p}_x$ of a texel $x$ is then defined as
\begin{equation}
	\mathbf{p}_x = f(K_x)
\end{equation}
and its \textit{argmax} defines the texel's class.

Previous work on mesh-based label fusion \citep{rosu2020semi} uses only the maximum of the probability distribution in the aggregator function due to performance reasons, as shown in \eqref{eq:maxsum}.

\begin{equation}
	\label{eq:maxsum}
	f_{\textit{maxsum}}(K) = \text{norm}\sum\limits_{k \in K} \begin{pmatrix}
		g_1(\mathbf{p}_k)\\
		\vdots\\
		g_c(\mathbf{p}_k)
	\end{pmatrix}
\end{equation}
\begin{equation*}
	\text{with } g_i(\mathbf{p}) = \begin{cases}
		\mathbf{p}_i, & \text{if }\mathbf{p}_i = \max \mathbf{p}\\
		0, & \text{if }\mathbf{p}_i < \max \mathbf{p}
	\end{cases}
\end{equation*}
\begin{equation*}
	\text{and } \text{norm}(\mathbf{p}) = \frac{\mathbf{p}}{\lVert \mathbf{p} \rVert_1}
\end{equation*}

Due to our optimized CUDA implementation and custom texture parametrization, we are able to exploit the entire \mbox{$c$-dimensional} probability distribution in the label fusion. This allows us to utilize Bayesian updating \citep{mccormac2017semanticfusion,rosinol2020kimera} in our aggregation function as shown in \eqref{eq:mul}, which is defined by an element-wise multiplication of probabilities.

\begin{equation}
	\label{eq:mul}
	f_{\textit{mul}}(K) = \text{norm}\prod\limits_{k \in K} \mathbf{p}_k = \text{norm}\prod\limits_{k \in K} \begin{pmatrix}
		\mathbf{p}_{k,1}\\
		\vdots\\
		\mathbf{p}_{k,c}
	\end{pmatrix}
\end{equation}

We also evaluate the average pooling aggregator \citep{tateno2017cnn} shown in \eqref{eq:sum}, which uses summation of probabilities similar to \citet{rosu2020semi}, but does not discard parts of the probability distribution.

\begin{equation}
	\label{eq:sum}
	f_{\textit{sum}}(K) = \text{norm}\sum\limits_{k \in K} \mathbf{p}_k = \text{norm}\sum\limits_{k \in K} \begin{pmatrix}
		\mathbf{p}_{k,1}\\
		\vdots\\
		\mathbf{p}_{k,c}
	\end{pmatrix}
\end{equation}

All aggregation results are normalized to a sum of $\lVert \mathbf{p} \rVert_1=1$ to represent a proper probability distribution.

Once the pixels of all frames have been aggregated the texels of the mesh store a consistent representation of the semantics of the environment. The mesh can then be rendered from the reconstructed camera poses to produce new consistent 2D annotation images corresponding to the original video frames.

Assuming non-erroneous pixel correspondences, the rendered annotation images will statistically have more accurate pixel labels than the original network prediction due to the probabilistic fusion. In practice, this fusion effect is weighed against the negative impact of erroneous pixel correspondences which stem from inaccurate mesh reconstructions and too coarse mesh primitives.

\subsubsection{Pixel Weighting}

Previous works \citep{mccormac2017semanticfusion,rosu2020semi} perform label fusion under the assumption that individual pixel predictions of the segmentation network are \textit{independent and identically distributed}~(\textit{i.i.d.}) observations of the underlying texel class. Each pixel is therefore given an equal weight towards the final label of its corresponding texel. This implicitly gives higher weight to frames where the texel occupies more pixels.

To compensate this effect, we propose a novel weighting scheme based on the assumption that individual images rather than pixels are \textit{i.i.d.} observations of the underlying texel class. Each image is therefore given an equal weight \wrt the final label of a texel. This reduces the implicit overweight of highly correlated pixel predictions and thereby improves the fused annotation. Our evaluation in \secref{sec:component-eval} supports this decision.

The distinction between \textit{i.i.d.} pixels and \textit{i.i.d} images can effectively be realized by extending \mbox{\eqref{eq:maxsum} - \eqref{eq:sum}} with a weight factor $w_k$ per pixel $k$ as shown in \mbox{\eqref{eq:w-maxsum} - \eqref{eq:w-sum}}.

\begin{equation}
	\label{eq:w-maxsum}
	f_{\textit{w-maxsum}}(K) = \text{norm}\sum\limits_{k \in K} w_k \begin{pmatrix}
		g_1(\mathbf{p}_k)\\
		\vdots\\
		g_c(\mathbf{p}_k)
	\end{pmatrix}
\end{equation}

\begin{equation}
	\label{eq:w-mul}
	f_{\textit{w-mul}}(K) = \text{norm}\prod\limits_{k \in K} \mathbf{p}_k^{w_k}
\end{equation}

\begin{equation}
	\label{eq:w-sum}
	f_{\textit{w-sum}}(K) = \text{norm}\sum\limits_{k \in K} w_k \mathbf{p}_k
\end{equation}

The aggregator function thus interprets pixel $k$ as having occurred $w_k$ many times. Weighting pixels equally is achieved with
\begin{equation}
	w_k = w_k^{(P)} = 1
\end{equation}
and weighting images equally is achieved with
\begin{equation}
	w_k = w_k^{(I)} = \frac{1}{|K_i \cap K_{x_k}|}
\end{equation}
where $K_i \subset K$ is the set of pixels in image $i$ with \mbox{$k \in K_i$}.

\subsection{Implementation}

Our framework is divided into a module for rendering the environment mesh and a module for the texel-wise aggregation of probability distributions.

\subsubsection{Renderer}

The renderer takes as input a triangle mesh and the extrinsic and intrinsic camera parameters of a given frame and projects all triangles onto the camera plane. To handle occlusion, we maintain depth information in a z-buffer.

For each pixel $k$ and its texture coordinates $(u_k, v_k)$ on the projected triangle $t$, we compute an identifier $\textit{id}_x \in \{0, \cdots, |R_t| - 1 \}$ for the corresponding texel $x$ as shown in \eqref{eq:idtexel}.

\begin{equation}
	\label{eq:idtexel}
	\textit{id}_x = \frac{\lfloor s \cdot u_k \rfloor^2 + \lfloor s \cdot u_k \rfloor}{2} + \lfloor s \cdot v_k \rfloor
\end{equation}

We store both a triangle identifier $\textit{id}_t$ and the texel identifier $\textit{id}_r$ per pixel. This rendered image of identifiers is passed to the aggregator module.

We employ CUDA's data parallelism over the set of triangles to speed up the rendering process.

\subsubsection{Aggregator}

\begin{figure}
	\hspace{2mm}\includegraphics[width=72mm]{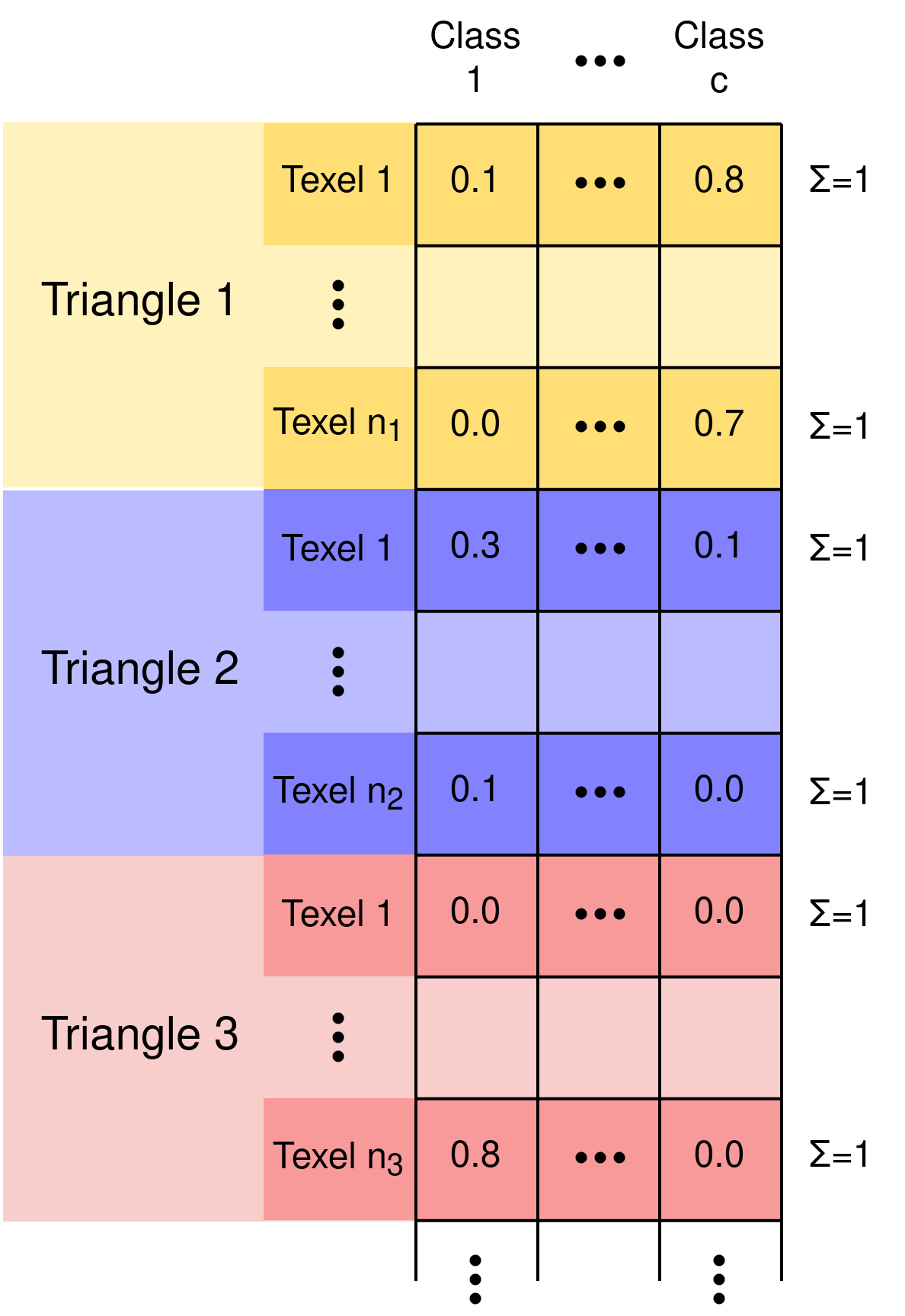}
	\caption{Layout of the two-dimensional probability distributions array $P$ with example probabilities. Every row describes a single texel. $n_i$ refers to the number of texels of the $i$-th triangle. Consecutive rows are grouped into triangles. Probabilities per row add up to $1$.}
	\label{fig:array-layout}
\end{figure}

The aggregator fuses the pixel predictions of all frames into a consistent representation on texel level.

We store the probability distributions of all texels in an array $P \in \mathbb{R}^{n_x \times c}$ where $n_x = \sum_{t \in T} |X_t|$ is the total number of texels in the mesh (\cf \figref{fig:array-layout}). The texels of all triangles are stacked along the first dimension of the array. Since triangles can have different number of texels, we define the triangle identifier $\textit{id}_t$ as its offset into the array along the first dimension for quick access. The pair of triangle and texel identifiers that are produced by the renderer module can therefore be used to find the corresponding row in $P$ as shown in \eqref{eq:texel-in-array}.

\begin{equation}
	\label{eq:texel-in-array}
	\mathbf{p}_x = P_l \text{ with } l = \textit{id}_{t_x} + \textit{id}_x
\end{equation}

For each frame, we first render the mesh and predict semantic pixelwise labels using the 2D segmentation model. For every pixel in the frame, the corresponding row in $P$ is then updated as defined by the aggregator function $f$ using the pixel's predicted probability distribution. The aggregator function $f$ is defined to be permutation-invariant to achieve deterministic results up to floating point inaccuracies.

The aggregator module can both be employed on the graphics processing unit (GPU) and the central processing unit (CPU). While the former avoids having to transfer the segmentation model's prediction from GPU to CPU and thus increases speed, the latter can accommodate larger meshes and higher texel resolutions which are more memory-intensive.

\section{EVALUATION}

\renewcommand\figwidth{0.48}
\renewcommand\figheight{50mm}
\begin{figure*}[t]
	\centering
	\begin{subfigure}[t]{\figwidth\textwidth}
		\centering
		\includegraphics[height=\figheight]{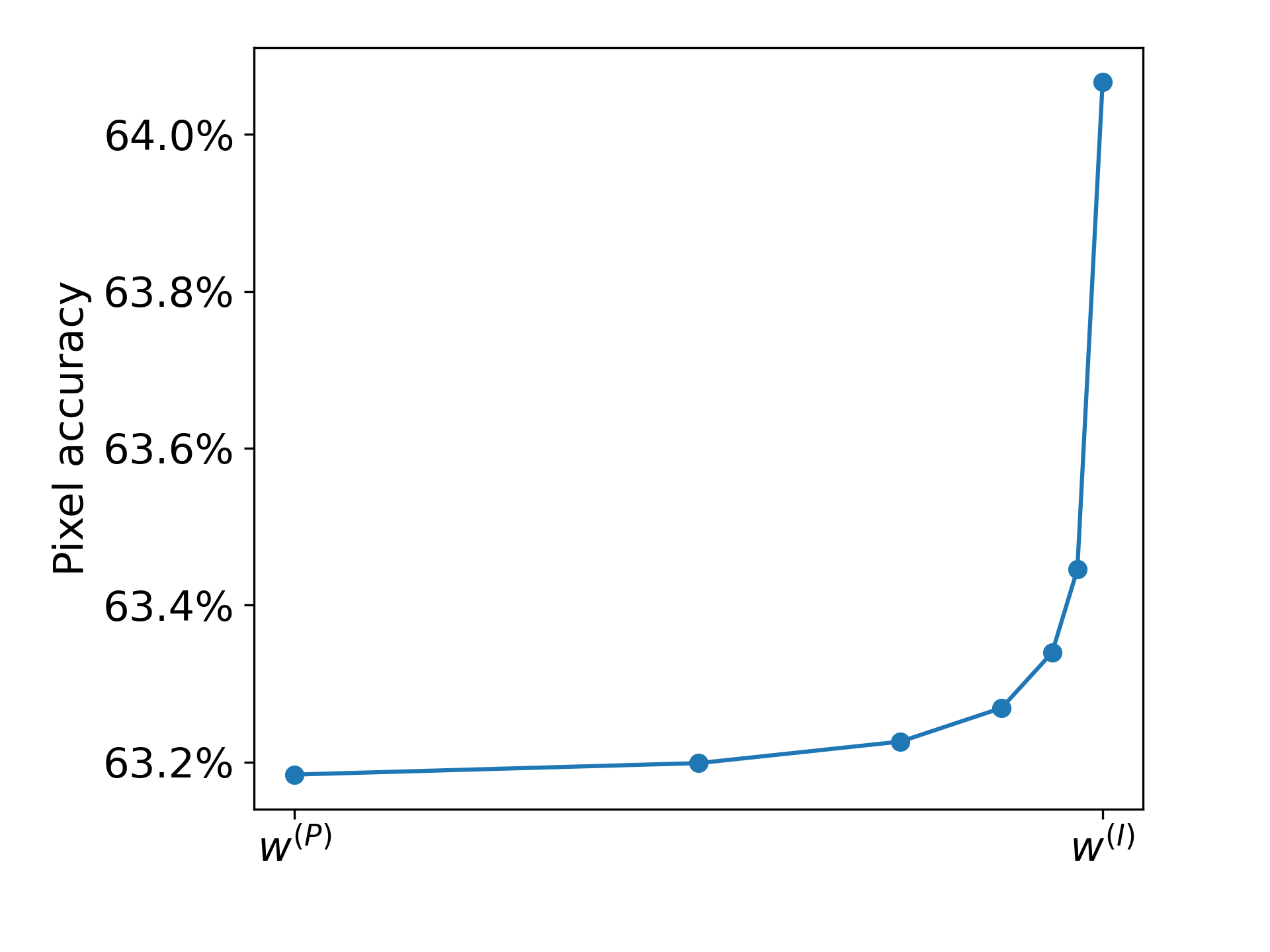}
		\vspace{-2mm}\caption{Pixel accuracy without sub-division of triangles (i.e. using a texel resolution $\gamma=0$) for different pixel weights \mbox{$w \in [w^{(P)}, w^{(I)}]$}. Values between $w^{(P)}$ and $w^{(I)}$ represent a weighted combination of both interpretations.}
		\label{fig:eval-imagesequalweight}
	\end{subfigure}\hfill%
	\begin{subfigure}[t]{\figwidth\textwidth}
		\centering
		\includegraphics[height=\figheight]{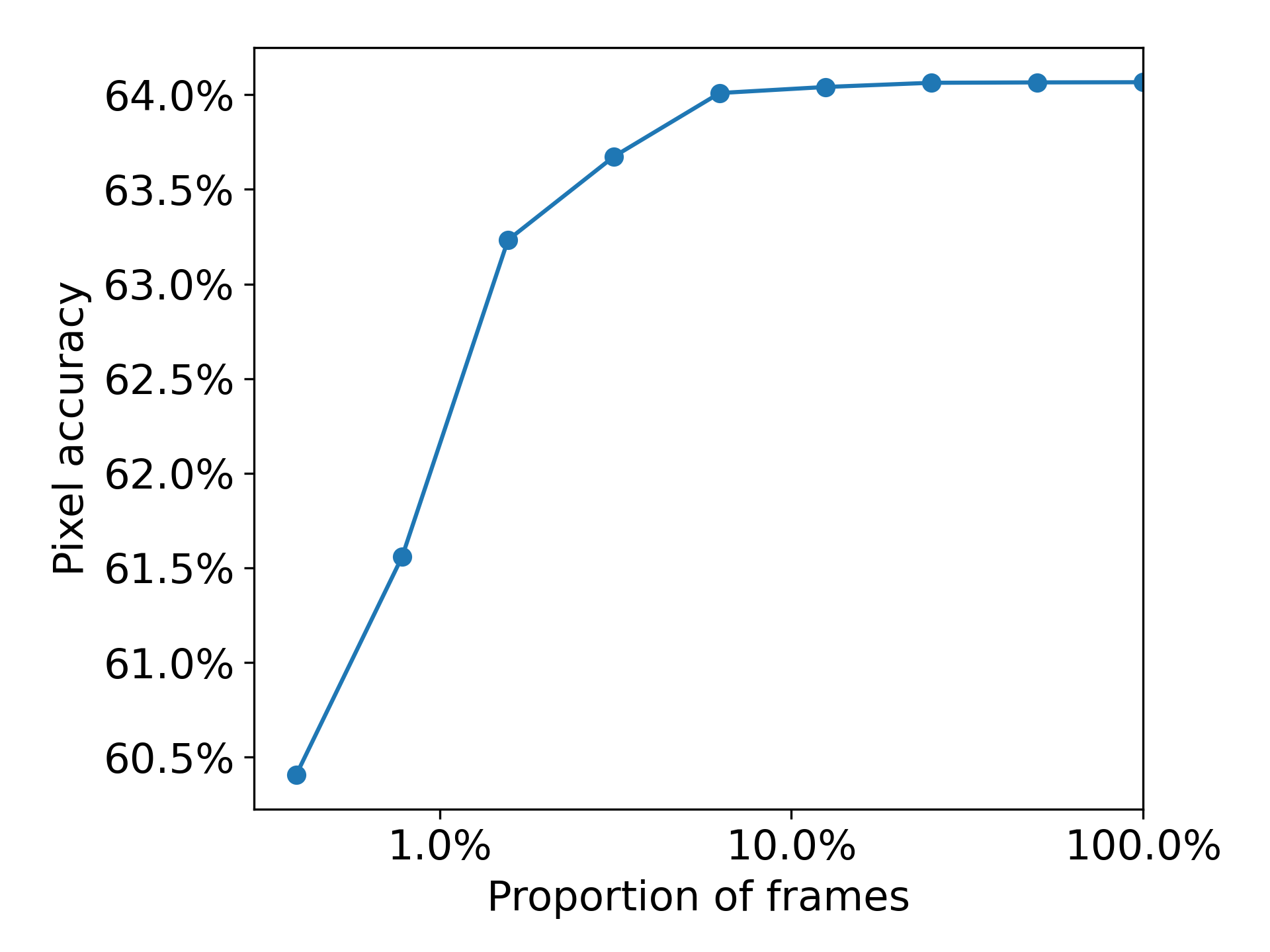}
		\vspace{-2mm}\caption{Pixel accuracy for different subsets of the available frames without sub-division of triangles (i.e. using a texel resolution $\gamma=0$) and with pixel weights $w^{(I)}$. Frames are chosen at uniform intervals.}
		\label{fig:eval-framesstep}
	\end{subfigure}
	
	\vspace{3mm}\begin{subfigure}[t]{\figwidth\textwidth}
		\centering
		\includegraphics[height=\figheight]{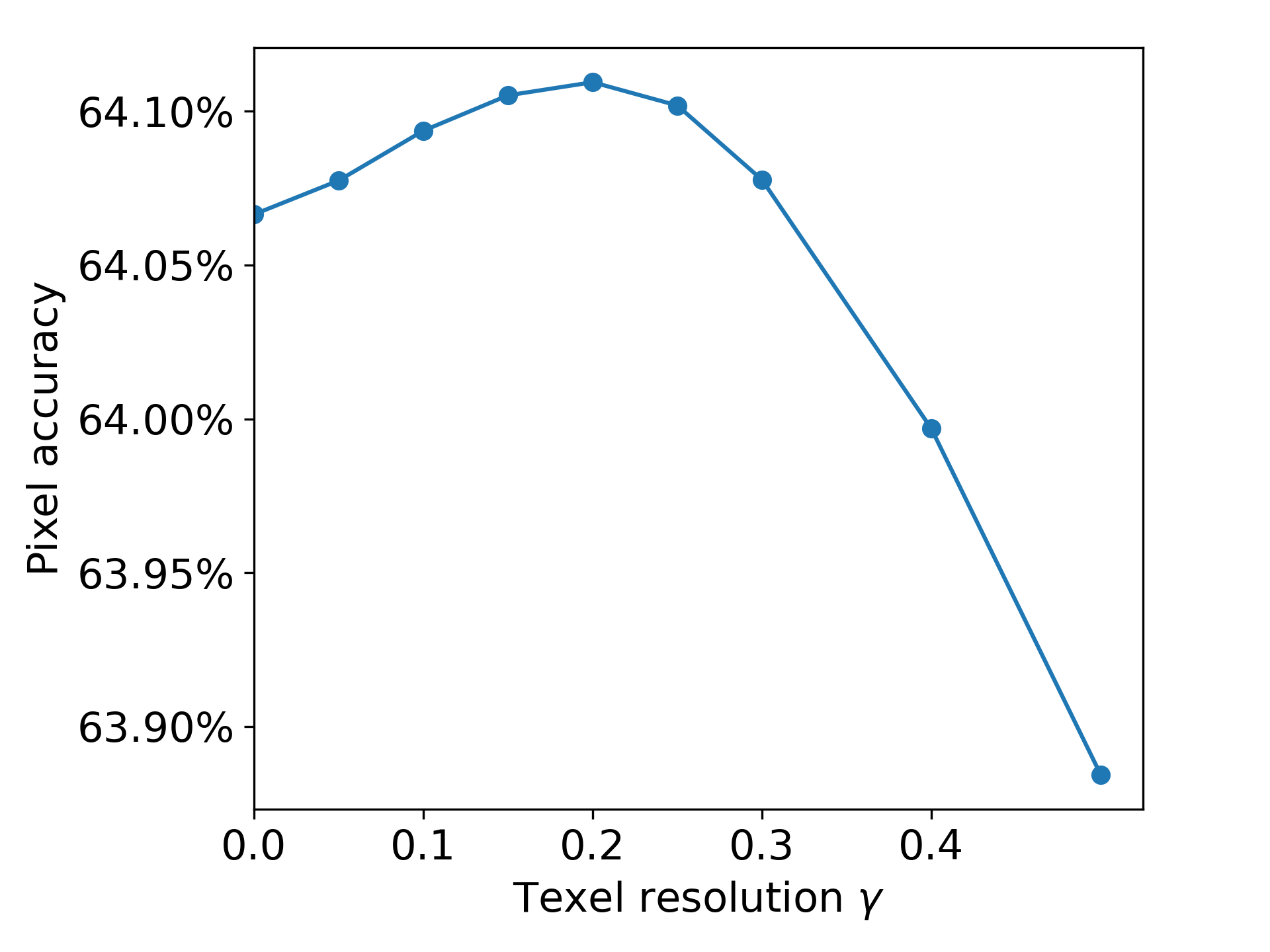}
		\vspace{-2mm}\caption{Pixel accuracy for different texel resolutions $\gamma$ with pixel weights $w^{(I)}$.}\label{fig:eval-texelres}
	\end{subfigure}\hfill%
	\begin{subfigure}[t]{\figwidth\textwidth}
		\centering
		\includegraphics[height=\figheight]{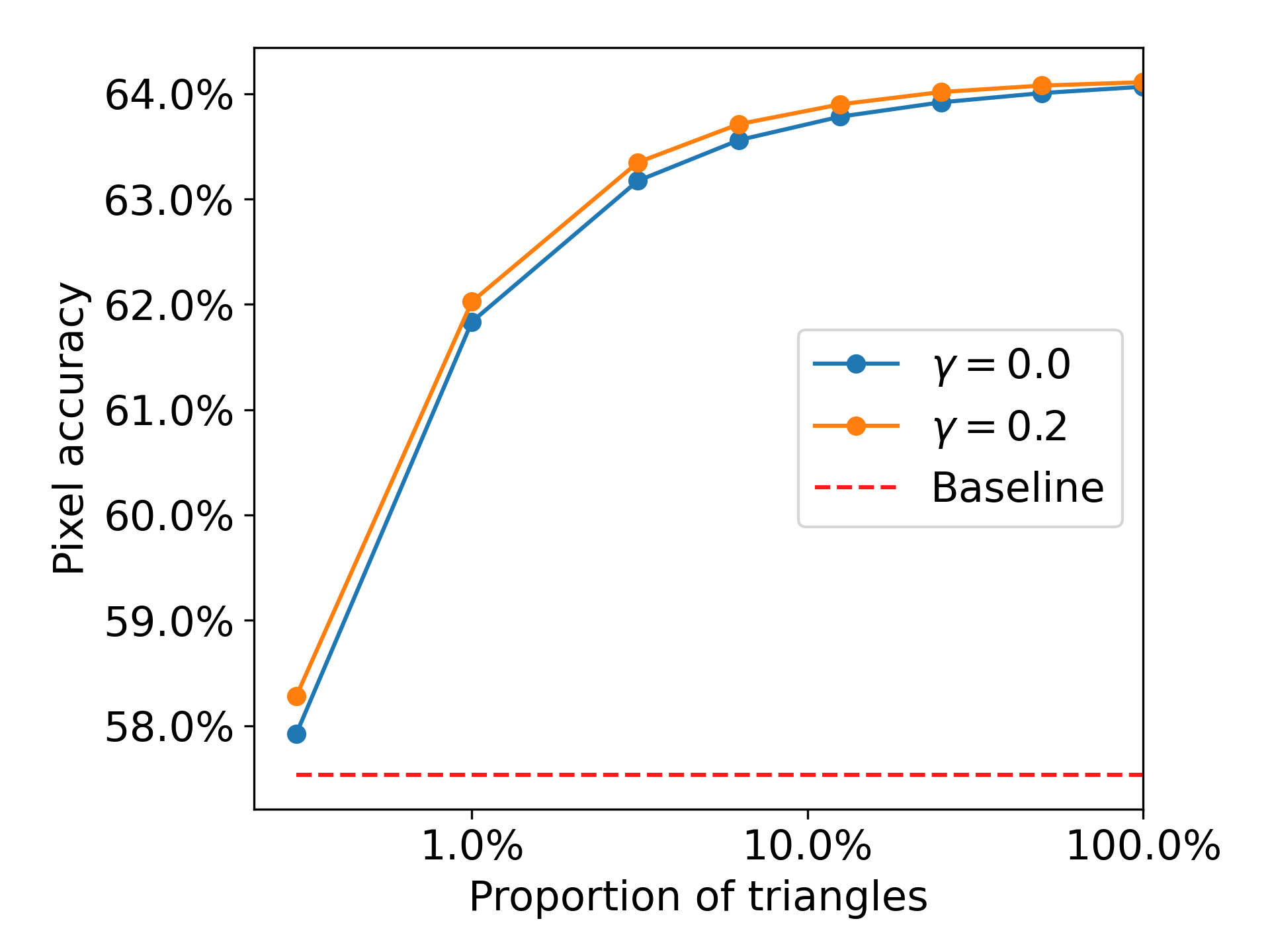}
		\vspace{-2mm}\caption{Pixel accuracy for meshes simplified to a proportion of triangles with pixel weights $w^{(I)}$. The value $\gamma = 0.2$ is chosen based on the results shown in \figref{fig:eval-texelres}.}\label{fig:eval-simplify}
	\end{subfigure}%
	\vspace{2mm}\caption{Analysis of factors that impact the label fusion using the first 100 scenes of the training split of Scannet. The graphs show the pixel accuracy after applying the label fusion step. The pixel weights $w^{(P)}$ and $w^{(I)}$ represent the assumptions of independent and identically distributed images and pixels, respectively. The aggregator function $f_{mul}$ is used in all experiments. Values between individual measurements are interpolated. \label{fig:eval-hyperparams}}
	\vspace{-0.3cm}
\end{figure*}

\subsection{Data and Architecture}
\label{sec:a2d2}

Evaluating our method requires densely labeled video sequences which only few publicly available datasets contain. This stems from the fact that redundant labels of subsequent frames represent a low cost-benefit ratio for other tasks like single image segmentation. We therefore evaluate our method on the Scannet v2 dataset \citep{dai2017scannet} of indoor scenes which contains densely labeled video data and corresponding meshes reconstructed with depth sensor data. We use the training split to evaluate the hyper-parameters of our method and report the final results on the validation split. We also create dense reconstructions of the first 20 scenes using Colmap and Delaunay triangulation \citep{schonberger2016pixelwise,labatut2009robust} to evaluate the performance on meshes created using a multi-view stereo approach.

For the semantic segmentation we choose a set of $40$ classes following the definition of the NYU Depth v2 dataset \citep{Silberman:ECCV12}. As segmentation model, we use ESANet \citep{DBLP:journals/corr/abs-2011-06961} trained on NYU Depth v2.

\subsection{Test Results}

We define our baseline as the original predictions of the segmentation network and use the validation split of the Scannet dataset for evaluation. This corresponds with a pixel accuracy of $52.05 \%$.

Based on the results of the extended evaluation in \secref{sec:component-eval}, we choose the aggregator function as $f_{mul}$, define pixel weights as $w^{(I)}$ and set the texture resolution to $\gamma = 0.2$. With this setup, our label fusion method improves the pixel accuracy on the validation split to $58.25 \%$.

\subsection{Extended Evaluation}
\label{sec:component-eval}

In the following, we examine the impact of several factors on the relative improvement of the label fusion over the original network prediction on the first 100 scenes of the training split of Scannet. The network achieves $57.53 \%$ pixel accuracy on this split. We use this evaluation to determine optimal hyper-parameters for the test results.

\textbf{Aggregator Function} \tabref{fig:eval-aggregators} shows the resulting pixel accuracy of different aggregator functions used in the label fusion. Bayesian fusion with $f_{mul}$ achieves the largest improvement, both with the \textit{i.i.d.} images and \textit{i.i.d.} pixels assumptions.

\renewcommand{\arraystretch}{1.4}
\begin{table}
	\centering
	\caption{Pixel accuracy of different aggregators in percent without sub-division of triangles (i.e. using the texel resolution $\gamma=0$). To the best of our knowledge, this is the first work to fuse full probability distributions (such as in $f_{mul}$ and $f_{sum}$) on semantic mesh textures.}
	\begin{tabular}{ c|c|c }
		Aggregator Function & $w^{(P)}$ & $w^{(I)}$
		\\
		\hline
		$f_{maxsum}$ \citep{rosu2020semi} & $62.85$ & $64.00$ \\
		$f_{sum}$ (ours) & $62.89$ & $64.04$ \\
		$f_{mul}$ (ours) & $63.18$ & $\textbf{64.07}$ \\
	\end{tabular}
	\label{fig:eval-aggregators}
\end{table}

\textbf{Pixel Weighting} Choosing $w^{(I)}$ as pixel weight shows improvements over $w^{(P)}$  for each aggregator function. Our measurements suggest that pixel accuracy increases monotonically for $w \in [w^{(P)}, w^{(I)}]$ which supports our decision to interpret the network predictions \textit{i.i.d.} images rather than \textit{i.i.d.} pixels (\cf \figref{fig:eval-imagesequalweight}).

\textbf{Frame Selection} \figref{fig:eval-framesstep} shows the resulting pixel accuracy when performing the label fusion with only a subset of the available frames chosen at uniform intervals. Using fewer frames also results in less information per texel that can be fused in a probabilistic manner, and thus decreases pixel accuracy. Above $20 \%$ of the frames we observe that additional images provide no significant improvement in pixel accuracy due to the high redundancy of information in adjacent frames.

\textbf{Texel Resolution} \figref{fig:eval-texelres} shows the fused pixel accuracy over different texel resolutions. Results improve slightly from $\gamma = 0.0$ to $\gamma = 0.2$ due to a finer texture resolution. For larger $\gamma$ values accuracy starts to decrease due to the smaller benefit of spatial consistency and stronger aliasing effects.

Using sub-triangle texture primitives with \mbox{$\gamma > 0$} improves accuracy by at most $0.03 \%$ compared to the original triangles as primitives with $\gamma=0$. This indicates that the Scannet meshes already have a sufficient granularity for the label fusion task. To test the performance on coarser meshes, we simplify the Scannet meshes using Quadric Edge Collapse Decimation \citep{meshlab}. This results in a smaller set of larger triangles that represent a suitable approximation of the original mesh geometry. We choose the best texel resolution (\ie \mbox{$\gamma = 0.2$}) based on the results shown in \figref{fig:eval-texelres} and compare with  the original triangles as primitives (\ie \mbox{$\gamma = 0$}). For a fixed $\gamma > 0$, the number of texels is roughly constant over different levels of simplification, while at $\gamma = 0$ the number of primitive elements decreases with the granularity of the mesh.

\figref{fig:eval-simplify} shows the pixel accuracy on meshes with different levels of simplification. The relative advantage of using sub-triangle textures increases for coarser meshes. When reducing the number of triangles to $0.3 \%$ of the input mesh, this results in a $0.3 \%$ absolute difference in pixel accuracy. For stronger simplifications, the geometric errors outweigh the benefits of the probabilistic fusion. This results in a lower pixel accuracy than the baseline annotations produced by the segmentation network.

\textbf{Input Mesh} Applying the label fusion on meshes of the first 20 scenes reconstructed with Colmap improves pixel accuracy from $57.46 \%$ to $59.67 \%$. This demonstrates that our method can also be applied on meshes that are created via a multi-view stereo approach.

\section{\uppercase{Conclusions}}
\label{sec:conclusion}

We have presented a label fusion framework that is capable of producing consistent semantic annotations for environment meshes using a set of annotated frames. In contrast to previous works, our method allows us to exploit the complete probability distributions of semantic image labels during the fusion process. We utilize the mesh representation as a long-term consistency constraint to also improve label accuracy in the original frames. We performed extensive evaluation of the proposed approach using the Scannet dataset, including the impact of factors like aggregation functions, pixel weighting, frame selection or texel resolution. Our experiments demonstrate that the proposed method yields a significant improvement in pixel label accuracy and can be used even with purely image-based multi-view stereo approaches. We make the code to our framework publicly available, including a CUDA implementation for efficient label fusion and a Python wrapper for easy integration with machine learning frameworks.

\bibliographystyle{apalike}
{\small
\bibliography{bib}}

\end{document}